\documentclass{article} 
\usepackage{amssymb}
\usepackage{amsmath}
\usepackage{nips14submit_e}
\usepackage{times}
\usepackage{hyperref}
\usepackage{url}
\usepackage{graphicx}
\usepackage{comment}

\DeclareMathOperator*{\argmin}{\mathrm{argmin}}
\DeclareMathOperator*{\minimize}{\mathrm{minimize}}

\DeclareMathOperator*{\diag}{\mathrm{diag}}
\DeclareMathOperator*{\vect}{\mathrm{vec}}

\title{Probabilistic Segmentation via Total Variation Regularization}

\author{
  Matt Wytock  \\
Machine Learning Department\\
Carnegie Mellon University \\
\texttt{mwytock@cs.cmu.edu} \\
\And
J. Zico Kolter \\
Machine Learning Department \\
Carnegie Mellon University \\
\texttt{zkolter@cs.cmu.edu}
}

\nipsfinalcopy 

\begin{document}

\maketitle

\begin{abstract}
We present a convex approach to probabilistic segmentation and
modeling of time series data.  Our approach builds upon recent
advances in multivariate total variation regularization, and seeks
to learn a separate set of parameters for the distribution over the
observations at each time point, but with an additional penalty that
encourages the parameters to remain constant over time.  We propose
efficient optimization methods for solving the resulting (large)
optimization problems, and a two-stage procedure for estimating
recurring clusters under such models, based upon kernel density
estimation.  Finally, we show on a number of real-world segmentation
tasks, the resulting methods often perform as well or better than
existing latent variable models, while being substantially easier to
train.
\end{abstract}

\section{Introduction}
In this paper, we consider the tasks of time series segmentation and
modeling.  Formally, suppose that we observe a sequence of $T$
input/output pairs, represented as
\begin{equation}
(x_1,y_1), (x_2, y_2), \ldots, (x_T, y_T)
\end{equation}
for $x_t \in \mathbb{R}^n$ (which can even include functions of past
outputs of the time series to capture scenarios such as
autoregressive models) and $y_t \in \mathbb{R}^p$ (though we can
also consider other possible forms of the output vector, such as
categorical variables).  Our goal is twofold: 1) to segment the time
series into (potentially non-contiguous) partitions $x_{\mathcal{I}_1},
\ldots, x_{\mathcal{I}_k}$, such that all the time points within each
partition can be modeled via a single probabilistic model $p(y_t |
x_t; \theta_i), \;\; \forall t \in \mathcal{I}_i$, parameterized by
$\theta_i \in \mathbb{R}^d$; and 2) to determine the
parameter,s $\theta_i$, of each different segment.  This is an extremely
general problem formulation and captures many of the aims of time
series latent variable models like hidden Markov models and their many
extensions \cite{rabiner1989tutorial}, multiple change-point detection
methods \cite{basseville1993detection}, and switched dynamical systems
\cite{sun2006switched}.

\begin{figure}[t]
\label{fig-model}
\centering
\includegraphics[width=2.4in]{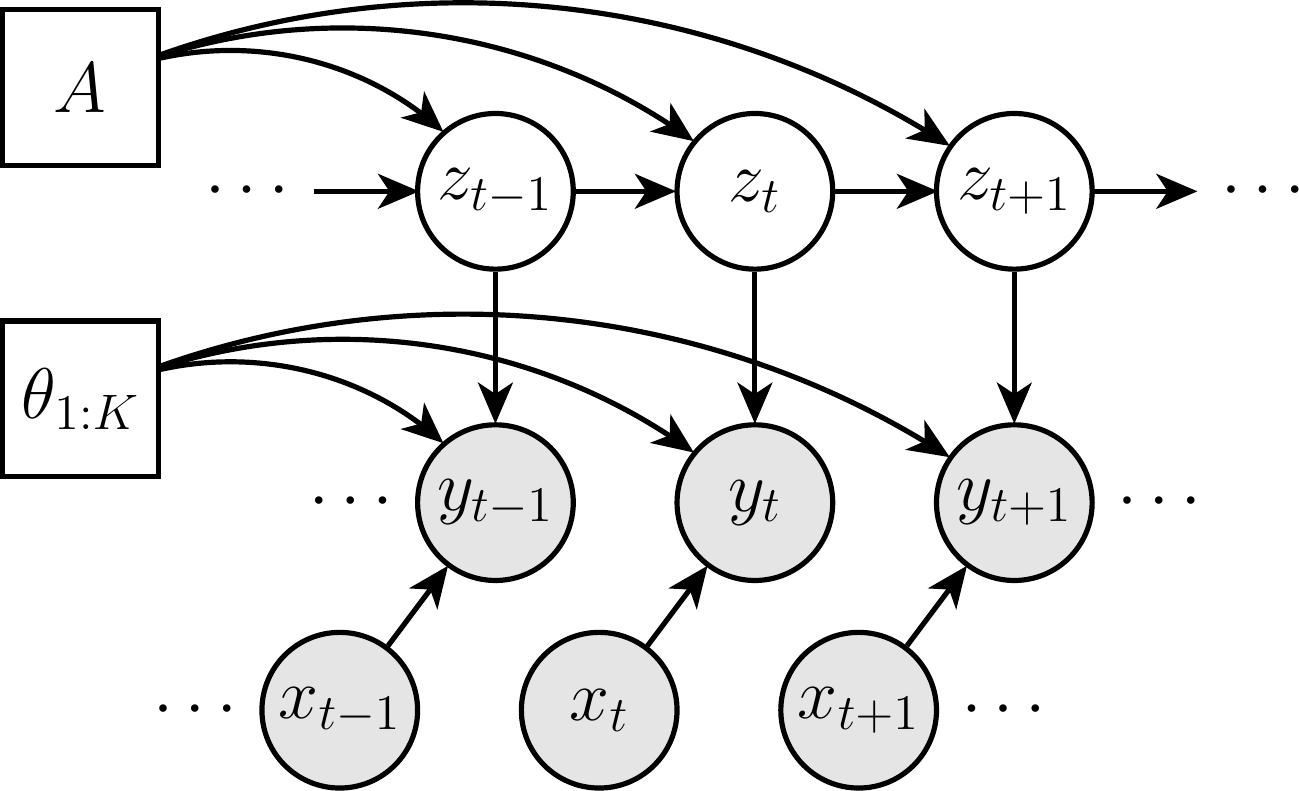}
\hspace{0.4in}
\includegraphics[width=2.4in]{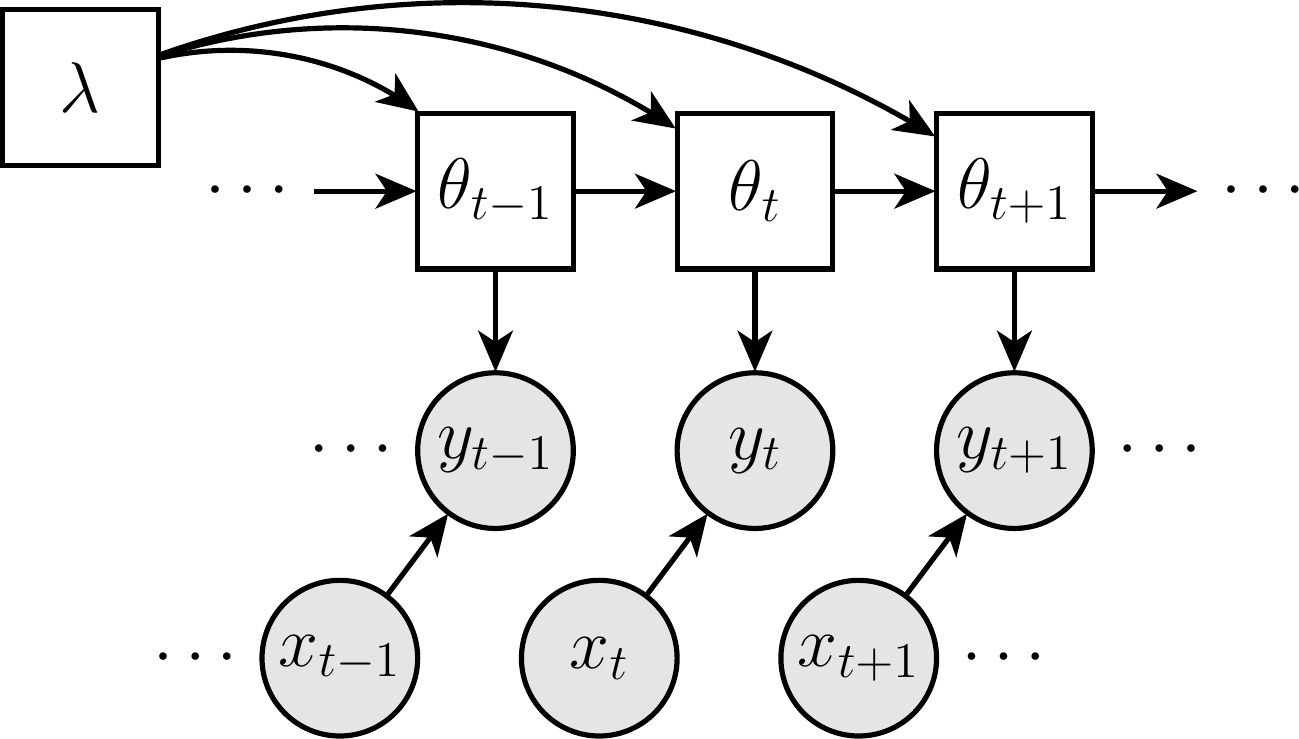}
\caption{A generic latent variable time series model (left) and our
  convex framework for segmentation and modeling (right).}
\end{figure}

A common approach to such tasks is what we generically refer to as a
latent state mixture model, with an illustrative example shown in
Figure \ref{fig-model} (left).  In such models, we associate with each
time a discrete latent state $z_t \in  \{1,\ldots,k\}$ that
``selects'' parameters to use for modeling the conditional
distribution $p(y_t  | x_t, z_t) = p(y_t | x_y; \theta_{z_t})$.  Under
such a model, the segmentation and  modeling challenges above can be
addressed respectively by e.g., 1) computing the most likely sequence
of latent states; and 2) jointly inferring the distribution over hidden
states and model parameters through the classical EM algorithm.
Though such models are very powerful, the fact that the EM algorithm
can be highly susceptible to local optima often makes learning such
models difficult, especially for complex distributions with many
parameters.  Furthermore, although the model over the latent
variables, $p(z_{t+1} | z_t; A)$ can capture complex dynamics in the
system, in practice a primary characteristic that these models must
capture is simply a  ``stickiness'' property: the fact that a system
in state $z_t$ tends to stay in this state with relatively high
probability; indeed, including such properties in latent variable  models has
been crucial to obtaining good performance \cite{fox2011sticky}.

In this work, we propose to capture very similar behavior, but in a
fully convex probabilistic framework.  In particular, we directly
associate with each time point a \emph{separate} set of parameters,
$\theta_t$, but then encourage the parameters to remain constant over
time by penalizing the $\ell_2$ norm of the difference between
successive parameters.  As is well-known from the group lasso setting
\cite{yuan2006model}, a penalty on the $\ell_2$ norm will encourage
group sparseness in the differences, i.e., a piecewise-constant
sequence of parameters; a graphical representation of this approach
is  show in Figure \ref{fig-model}.  Formally, we jointly segment and
model the system by solving the (convex) optimization problem
\begin{equation}
  \label{eq-prob}
  \minimize_{\theta_1,\ldots,\theta_T} \;\; -\sum_{t=1}^T \log
  p(y_t|x_t;\theta_t) +  \lambda \sum_{t=1}^{T-1}\|\theta_{t+1} -
  \theta_t\|_2.
\end{equation}
This penalty function on $\theta$ is sometimes referred to
as the group fused lasso \cite{alaiz2013group,bleakley2011group} or
the multivariate total variation penalty, but the key of our approach
is to apply such penalties to the underlying parameters of the
probability distribution rather than to the output signal itself.
The resulting model generalizes several existing
methods from the literature, including the standard group fused lasso
itself \cite{bleakley2011group}, time-varying linear regression
\cite{ohlsson2013identification} and auto-regressive modeling
\cite{ohlsson2010segmentation}, and $\ell_1$ mean and variance
filtering \cite{wahlberg2012admm}.

Although the proposed approach is conceptually simple, there are a
number of elements needed to make this approach practical;
together, these make up the majority of our contribution in this work.
First, the optimization problem
\eqref{eq-prob} is challenging: it involves a $kT$ dimensional optimization
variable, potentially non-quadratic loss functions, a non-smooth
$\ell_2$  norm penalty, and achieving \emph{exact} sparseness in these
differences is crucial given that we want to use the method to split
the time series into distinct segments.  Second, a
major advantage of latent variable models is that they can capture
disjoint segments, where the underlying parameters change and return
to previous values; this structure is inherently missed by the total
variation penalties, as there is no innate mechanism by which we can
``return'' to previous parameter values.  Instead, we propose to
capture much of this same behavior via a two-pass algorithm that
clusters the parameters themselves using kernel density estimation.
Finally, the main message of this paper is an empirical one, that the
convex framework \eqref{eq-prob} can perform as well, in practice, as
much more complex latent variable models, while simultaneously being
much faster and easier to optimize.  Thus, we present empirical
results on three real-world domains studied in the latent variable
modeling literature: segmenting honey bee motion, detecting and
modeling device energy consumption in a home, and segmenting motion
capture data.  Together, these illustrate the wide applicability of
the proposed approach.

\section{Efficient ADMM optimization method with fast Newton subroutines}

In this section, we develop fast algorithms for solving the probabilistic
segmentation problem \eqref{eq-prob} with different probabilistic
models $p(y_t | x_t;\theta_t)$, a necessary step for applying the model to
real data sets. Although the problem is convex, optimization is
complicated by the non-smooth nature of the total variation norm (i.e. exactly
the structure that promotes sparsity in the change points which we desire) and
the composite objective incorporating a probability distribution with a possibly
different set of parameters at each time point. The result is a
difficult-to-solve optimization problem, and we found that
off-the-shelf solvers performed poorly on even moderately sized
examples. Our approach to
optimization decomposes the objective into many smaller subproblems
using the alternating direction methods of multipliers (ADMM)
\cite{boyd2011distributed}---iterating between solving the subproblems
and taking a gradient step in the dual.


Omitting details for the sake of brevity (the derivations here are straightforward and for a thorough
description of ADMM including complete description of the algorithm, we refer readers to \cite{boyd2011distributed} and a
similar form form of ADMM, though just for quadratic loss, is also described in
\cite{wytock2014fast}), for problems of the form \eqref{eq-prob}, the
algorithm iteratively performs the following updates starting at some
initial $\theta_{1:T}^0$, $z_{1:T}^0$, and $u_{1:T}^0$:
\begin{equation}
\label{eq-admm}
\begin{split}
\theta^{k+1}_t & \leftarrow \argmin_{\theta_t} \;\; -\log p(y_t | x_t ;
\theta_t) + \frac{\rho}{2} \|\theta_t - z^k_t + u^k_t\|_2^2, \;\; \forall
t=1,\ldots,T \\
z^{k+1}_{1:T} & \leftarrow \argmin_{z_{1:T}} \;\; \sum_{t=1}^{T-1} \|z_t -
z_{t+1}\|_2 + \frac{\rho}{2} \sum_{t=1}^T \|x^{k+1}_t - z_t + u^k_t\|_2^2 \\
u^{k+1}_{t} & \leftarrow u^k_t + x^{k+1}_t - z^{k+1}_t, \;\; \forall
t=1,\ldots,T.
\end{split}
\end{equation}
ADMM is particularly appealing for such problems because the $z$
update here is precisely a group fused lasso, a problem for
which  efficient second-order methods exist \cite{wytock2014fast}, and
because the $\theta_t$ updates are separate, which allows the method
to be trivially parallelized. In addition,
the proposed ADMM approach is appealing because it is extensible---for
example, we could encode
additional structure in the problem by penalizing the trace norm
\cite{recht2010guaranteed} of $[\theta_1,\ldots,\theta_T]$, an extension that is
straightforward requiring only minor modifications to the algorithm
and the implementation of the proximal operator for this new penalty
(in this case, thresholding on singular values).

While fast algorithms exist for the total variation norm, the novel element
required for our problem is efficient implementation of the proximal operators
for the log-loss term, that is, the $\theta$ updates in
\eqref{eq-admm}. In particular (observing that the this term separates
over time points, we drop the subscript $t$) we derive efficient implementations
for the subproblems
\begin{equation}
  \minimize_\theta -\log p(y|x;\theta) + \frac{\rho}{2}\|\theta -
  \theta_0\|_2^2.
\end{equation}
Note that since our framework allows
for the possibility of different parameters at each time point, in each
iteration of ADMM we must solve this problem $T$ times resulting in different
estimates for $\theta_t$, a setting somewhat different than standard maximum
likelihood estimation in which we estimate one  set of parameters for many data
points. Furthermore, as we highlight in the sequel, minimizing the log-loss term
over only a single observation gives rise to additional structure which we can
exploit. Next, we consider fast updates for the cause of a
multivariate conditional Gaussian (with unknown covariance/precision
matrix), a natural distribution for our model.  The models are of
course also extensible to other probability models such as the softmax
model for discrete outputs, and the resulting method is very similar
to that presented below.  Code for the full algorithm will be
included with the final version of the paper.

\subsection{Gaussian model}

Suppose we have a continuous output variable $y \in
\mathbb{R}^p$ and we model $y|x$ as
\begin{equation}
p(y|x;\Lambda,\Theta) = \frac{1}{Z(x)}\exp\left(-\frac{1}{2}y^T\Lambda y -
x^T\Theta y \right)
\end{equation}
with parameters $\Lambda \in \mathbb{R}^{p \times p}$ and $\Theta \in
\mathbb{R}^{n \times p}$; note that under this model, $\Theta$ and
$\Lambda$ take the place of $\theta$ with $\theta = [\vect(\Lambda)^T,
\vect(\Theta)^T]^T$ and for the rest of this section we simply consider the
  parameters to be $\Lambda$ and $\Theta$ for ease of notation. This model is
  equivalent to  a multivariate Gaussian with $y|x \sim
\mathcal{N}(-\Lambda^{-1}\Theta^Tx,\Lambda^{-1})$ but this particular
parameterization with the scaled mean parameter is attractive as it admits
a convex regularized maximum likelihood estimation problem:
\begin{equation}
\minimize_{\Lambda,\Theta}  -\log \det \Lambda +
y^T\Lambda y + x^T\Theta^T\Lambda^{-1}\Theta x + 2y^T\Theta x +
\frac{\rho}{2} \left(\|\Lambda-\Lambda_0\|_F^2 + \|\Theta-\Theta_0\|_F^2  \right).
\end{equation}
This problem is convex and without the addition of the augmented
Lagrangian terms can be solved in closed form; however, with those terms no such
closed form exists and thus our approach is to develop a second-order Newton
method. We start by taking the gradient
\begin{equation}
\begin{split}
\nabla_\Lambda &=  - \Lambda^{-1} + yy^T -
\Lambda^{-1}\Theta^Txx^T\Theta\Lambda^{-1} + \rho(\Lambda - \Lambda_0)\\
\nabla_\Theta  &= 2xx^T\Theta\Lambda^{-1} + 2xy^T + \rho(\Theta-\Theta_0)
\end{split}
\end{equation}
and we solve for the Newton direction, parameterized by $(U,V)$ where $U$
represents the change in $\Lambda$ and $V$ in $\Theta$, by considering the
system of equations
\begin{equation}
\begin{split}
\Lambda^{-1}U\Lambda^{-1} +
\Lambda^{-1}\Theta^Txx^T\Theta\Lambda^{-1}U\Lambda^{-1} +
\Lambda^{-1}U\Lambda^{-1}\Theta^Txx^T\Theta\Lambda^{-1} \\
- \Lambda^{-1}V^Txx^T\Theta\Lambda^{-1}
- \Lambda^{-1}\Theta^Txx^TV\Lambda^{-1}
+ \rho U &= \nabla_\Lambda \\
2xx^TV\Lambda^{-1} - 2xx^T\Theta\Lambda^{-1}U\Lambda^{-1} + \rho V &= \nabla_\Theta
\end{split}
\end{equation}
which is a Sylvester-like equation that we could solve using the identity
$\vect(AXB) = (B^T \otimes A)\vect(A)$ where $\otimes$ denotes the Kronecker
product.

However, naively employing this approach requires inverting a $p(n+p)
\times p(n+p)$ matrix and thus is not computationally tractable for
reasonably sized problems. Instead, we simplify the system of
equations analytically so that solving for the Newton direction
requires only $O(p^3)$ operations. We proceed by taking the
eigendecomposition $\Lambda^{-1} = WSW^T$ and writing this system of
equations as
\begin{equation}
\begin{split}
S\tilde{U}S + S\tilde{\Theta}^Txx^T\tilde{\Theta}S\tilde{U}S +
S\tilde{U}S\tilde{\Theta}^Txx^T\tilde{\Theta}S \\
-S\tilde{V}^Txx^T\tilde{\Theta}S - S\tilde{\Theta}^Txx^T\tilde{V}S + \rho\tilde{U} &= W^T\nabla_\Lambda W \\
2xx^T\tilde{V}S -2xx^T\tilde{\Theta}S\tilde{U}S + \rho\tilde{V} &=
\nabla_\Theta W
\end{split}
\end{equation}
where $\tilde{U} = W^TUW$, $\tilde{V} = VW$ and $\tilde{\Theta} = \Theta
W$. Now using the $\vect$ operator we have
\begin{equation}
\label{eq-kron-newton}
\left[ \begin{array}{cc}
S \otimes S + S \otimes aa^T + aa^T \otimes S + \rho I & S \otimes
ax^T + (S \otimes ax^T)K_{np} \\
 S \otimes xa^T + K_{pn}(xa^T \otimes S) & 2S \otimes xx^T + \rho I
\end{array} \right]
\left[
 \begin{array}{cc}
\vect \tilde{U} \\  \vect \tilde{V}
\end{array}  \right] =
\left[
 \begin{array}{cc}
\vect{W^T\nabla_\Lambda W} \\ \vect \nabla_\Theta W
\end{array}  \right]
\end{equation}
where $a = -S\tilde{\Theta}^Tx$ and $K_{np}$ is the commutation matrix
(see e.g. \cite{magnus1988matrix}). Although the matrix on the LHS of this linear system is
large, it is highly structured; specifically it can be factorized into
diagonal and low rank components, written as $D + AA^T$ with
\begin{align}
D = \left[ \begin{array}{cc}
S \otimes S + \rho I & 0 \\
0 & \rho I
\end{array} \right]
&&
A = \left[ \begin{array}{cc}
S^{1/2} \otimes a & a \otimes S^{1/2} \\
S^{1/2} \otimes x & S^{1/2} \otimes x
\end{array} \right].
\end{align}
Next, using the matrix inversion lemma we have
\begin{equation}
\label{eq-matrix-inversion}
(D + AA^T)^{-1} = D^{-1} - D^{-1}A(I + A^TD^{-1}A)^{-1}A^TD^{-1}
\end{equation}
and after a bit of algebra, we observe that
\begin{equation}
\label{eq-mat-inv}
(I + A^TD^{-1}A) = \left[ \begin{array}{cc}
      X + C_1 & X + C_2 \\
      X + C_2 & X + C_1
\end{array} \right]
\end{equation}
where
\begin{align}
X = \frac{1}{\rho}(x^Tx)S, && C_1 = \diag\left( \frac{(a \circ a)s^T}{ss^T +
  \rho} 1 \right), && C_2 = \frac{S^{1/2}aa^TS^{1/2}}{ss^T + \rho}.
\end{align}
Using this form, we are able to compute each term in the Newton direction
$(D+AA^T)^{-1}\vect(G)$ where $G$ is the RHS of the equation from
\eqref{eq-kron-newton} without ever forming the Kronecker products
explicitly resulting in a computation complexity of $O(p^3)$, the cost to invert
\eqref{eq-mat-inv}.

\section{Segment clustering via kernel density estimation}

As mentioned above, a notable disadvantage our proposed convex
segmentation methods is that, unlike latent variable models, there is
no inherent notion of parameters being tied across disjoint segments
of the time series.  Indeed, the effect of the above segmentation
model will be to determine the best single model for each segment
individually (modulo the regularization penalties).  Although we
observe that, in real-world settings, this does not appear to be as
large a problem as might be imagined for learning the individual model
parameters themselves (the ``stickiness'' component mentioned above
typically means that there is enough data per segment to learn
good models), it is a substantial concern if the overall goal is to
make joint inferences about the nature of related segments in the same
time series or across multiple time series.  To this end, we advocate
for a two-stage alternative to the latent variable model:
in the first stage,we compute the convex segmentation as above and in
the second stage, we cluster the segments directly in parameter space,
via kernel density estimation.

While there are many methods to cluster points in Euclidean space, density
clustering using the kernel estimator is appealing as identifying modes in the
distribution over the parameter space fits well with our probabilistic model.
The intuitive idea is that given the true probability distribution over the parameter
space $p$ and a point $\theta \in \mathbb{R}^d$, we define the cluster for $\theta$
to be the mode found by following the gradient $\nabla p(\theta)$. In practice,
since we do not know the true underlying distribution, we replace $p$ with
$\hat{p}_h$, the kernel density estimator constructed with bandwidth $h$
\begin{equation}
\hat{p}_h(\theta) = \frac{1}{T} \sum_{t=1}^T \frac{1}{h^d} K \left(\frac{\|\theta-\theta_t\|}{h}\right)
\end{equation}
where $K$ is a smooth, symmetric kernel. The standard kernel choice (which we use) is the Gaussian
kernel; in this case, it is known that the number of modes of $\hat{p}_h$ is a
nonincreasing function of $h$ \cite{silverman1981using} and thus the clustering
is well-behaved. Furthermore, given this property, in practice we typically fix the number
of clusters based on the application and choose $h$ such that
kernel density clustering results in the desired number of
modes.  However, we note that there are several possibilities for a more nuanced
selection of the bandwidth---for example, we could select $h$ based on the terms
of the objective function or the difference in norms between adjacent segments.

\section{Experimental results}

In this section, we evaluate the proposed method on several applications,
some of which were previously considered in the context of parametric and
nonparametric latent variable models using Bayesian inference
\cite{fox2011bayesian,fox2013joint,oh2008learning,xuan2007modeling}. In these
applications, we typically demonstrate equal or better performance with a
substantially different approach---unlike the latent variables models, our
method is fully convex and thus not subject to local optima. Following the
direction of previous work, we treat these tasks as unsupervised with the
parameter $\lambda$ controlling the trade-off between the complexity of
the model (number of change points) and the data
fit. In addition to considering our probabilistic method  (``TV Gaussian''),
we also consider the previously proposed linear regression model
\cite{ohlsson2013identification} (``TV LR'') in which case the log-loss term
simply becomes the least squares penalty.

\begin{table}
\caption{Performance on dancing honey bees data set.}
\begin{center}
\begin{tabular}{|l|c|c|c|c|c|c|c|}
\hline & 1 & 2 & 3 & 4 & 5 & 6 & Average \\
\hline HDP-VAR(1)-HMM unsupervised & 46.5 & 44.1 & 45.6 & 83.2 & 93.2 & 88.7 &
66.9 \\
HDP-VAR(1)-HMM partially supervised  & 65.9 & 88.5 & 79.2 & 86.9 & 92.3 & 89.1 &
83.7 \\
SLDS DD-MCMC & 74.0 & 86.1 & 81.3 & 93.4 & 90.2 & 90.4 & 85.9 \\
PS-SLDS DD-MCMC & 75.9 & 92.4 & 83.1 & 93.4 & 90.4 & 91.0 & 87.7 \\
TV Linear regression  & 54.4 & 47.7 & 79.6 & 78.8 & 76.1 & 75.5 & 68.9 \\
TV Gaussian  & 82.2 & 83.3 & 76.1 & 91.1 & 93.1 & 93.1 & 86.5 \\
\hline
\end{tabular}
\label{tab-bees}
\end{center}
\end{table}

\begin{figure}[t]
\includegraphics{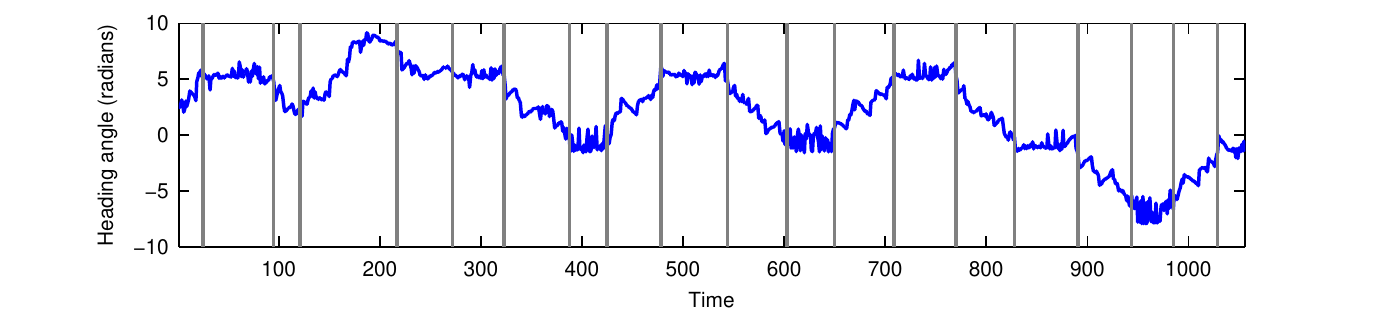}
\includegraphics{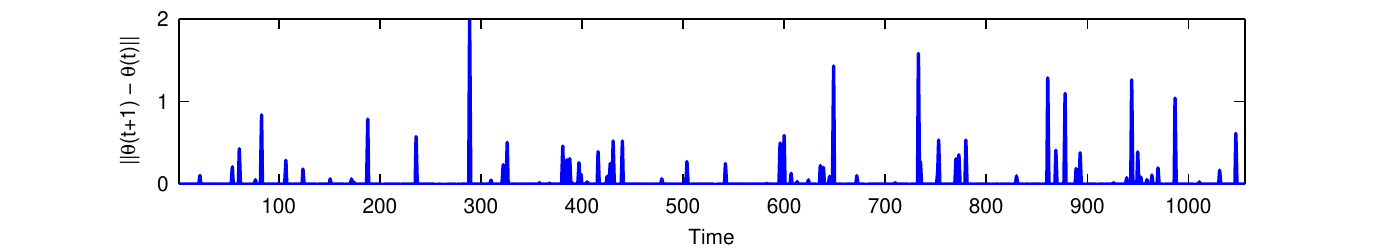}
\includegraphics{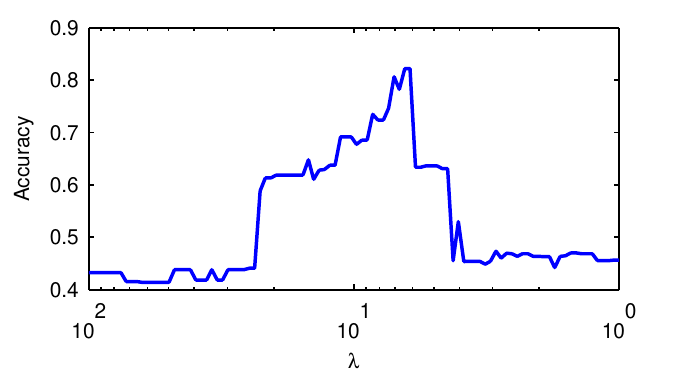}
\includegraphics[width=2.73in]{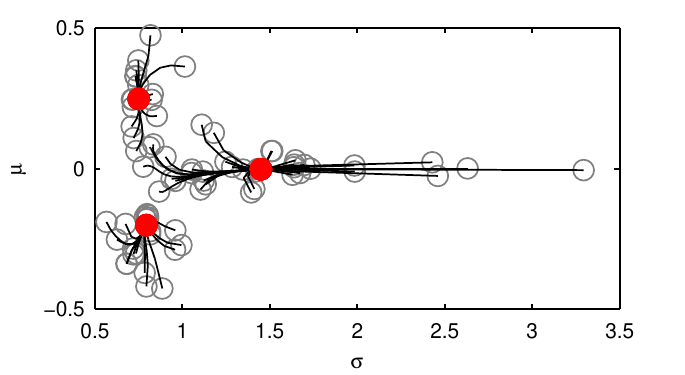}
\caption{Segmentation results for honey bee sequence 1 with top plot showing
  the observed heading angle of the bee along with the (unobserved) segmentation
  provided by human experts. In middle, we show the segmentation provided by the
  best choice of $\lambda$ and on bottom left the comparison of performance
  vs. $\lambda$. On bottom right, the parameters for each segment (gray circles)
  are clustered using kernel density estimation to identify three modes (red
  circles).}
\label{fig-bees}
\end{figure}

{\bf Dancing honey bees.} Our first data set involves tracking honey bees from
video, a task that was first considered in \cite{oh2008learning} and
subsequently studied by \cite{xuan2007modeling,fox2011bayesian}. The data includes
the output of a vision system which identifies the bees position and heading
angle over time and the task is to segment these observed values into three
distinct actions: turn left, turn right and ``waggle''---characterized by rapid
back and forth movement. It is known that these actions are used by the bees
to communicate about food sources, and for the 6 sequences provided we also have a ground truth
labeling of actions by human experts. In Figure \ref{fig-bees} (top) we show the
angle variable along with labels showing behavior changes; as can be seen from
the graph, we found that the change in angle to be highly indicative of the bees
behavior, to the extent that we model this time series ignoring the other
data. Specifically we take first order differences  $y_t = \phi_{t+1} - \phi_t$
and represent this sequence probabilistically as $y_t \sim
\mathcal{N}(\mu_t,\sigma^2_t)$, expecting that both the mean and variance to
change based on the bee's action.

In Table \ref{tab-bees} we compare the accuracy of our segmentation using the
best setting of $\lambda$ to that of previous work on this data set
\cite{oh2008learning,fox2011bayesian}. Although this is an optimistic
comparison, we observe that our method is not especially sensitive to $\lambda$
as can be seen in Figure \ref{fig-bees} (bottom left); in particular, while the
performance is particularly good for the best $\lambda$, there is a wide
range in which the model performs as well or better than the other
methods. It should also be noted that with the exception of ``HDP-VAR(1)-HMM
unsupervised'' all of the considered approaches include some level of
supervision (e.g. first training on the other 5 sequences) while our method
is fully unsupervised with only a single tuning parameter. Next, considering the
distribution of the parameters using kernel density estimation, we see in
Figure \ref{fig-bees} (bottom right) that our  method indeed identifies the 3
modes of the distribution corresponding to labeled actions: turning left/right
correspond to positive/negative mean while waggle has zero mean but
significantly larger variance. The change in variance offers one intuitive
explanation as to why the probabilistic model outperforms linear regression
on this data set since the latter does not model this behavior.

\begin{figure}[t]
 \includegraphics[width=2.73in,height=1in]{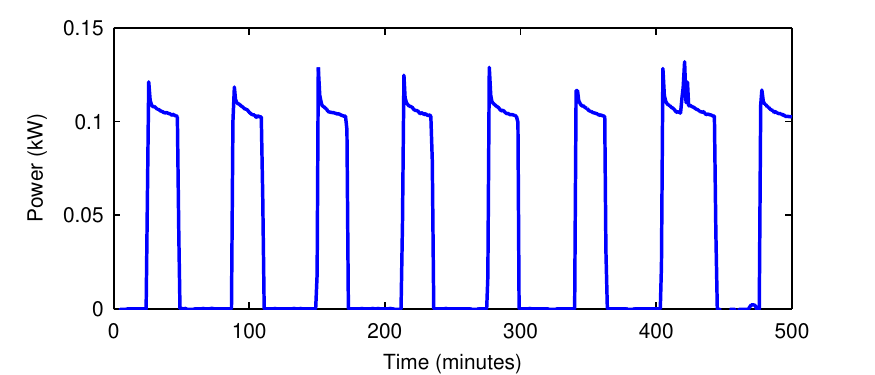}
 \includegraphics[width=2.73in,height=1in]{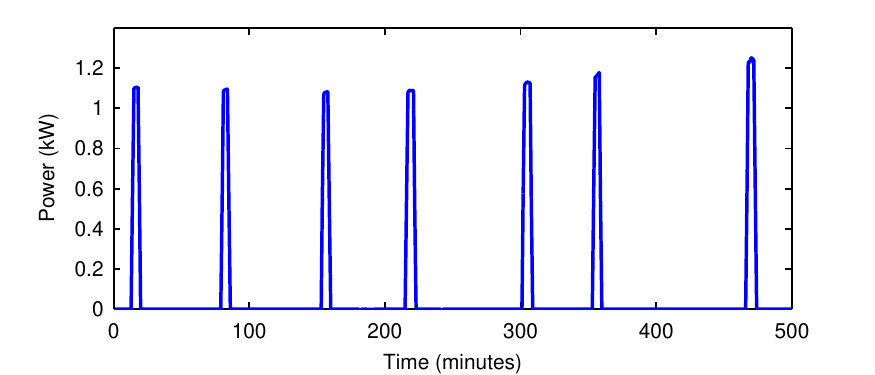}
 \includegraphics[width=2.73in,height=1in]{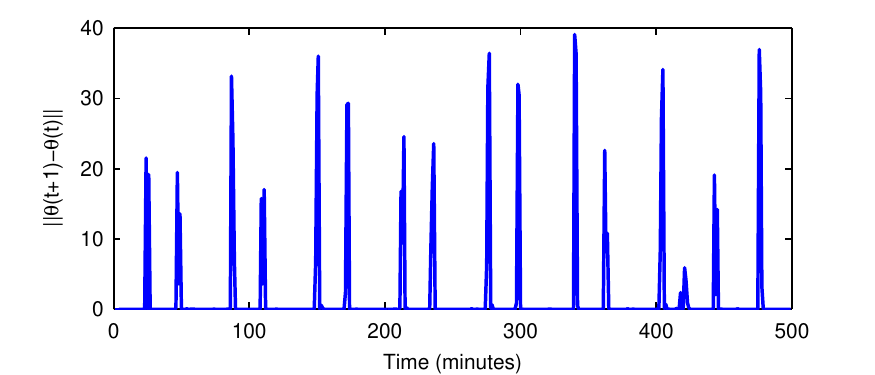}
 \includegraphics[width=2.73in,height=1in]{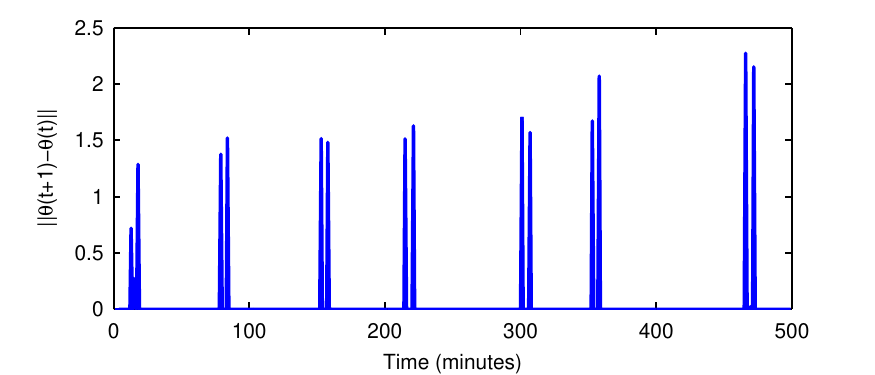}
 \includegraphics[width=2.73in,height=1in]{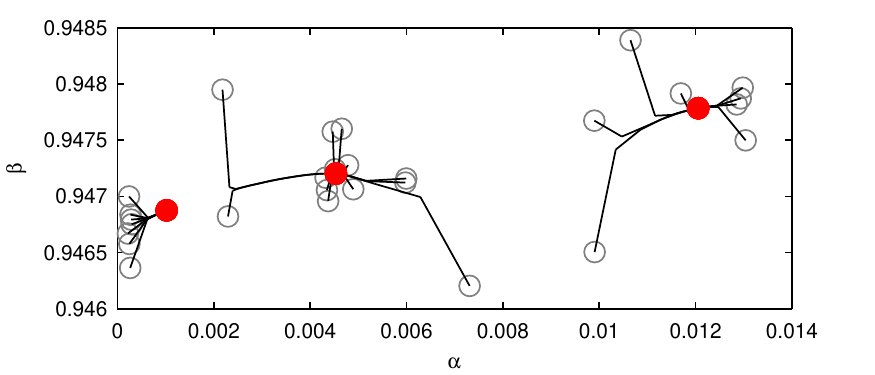}
 \includegraphics[width=2.73in,height=1in]{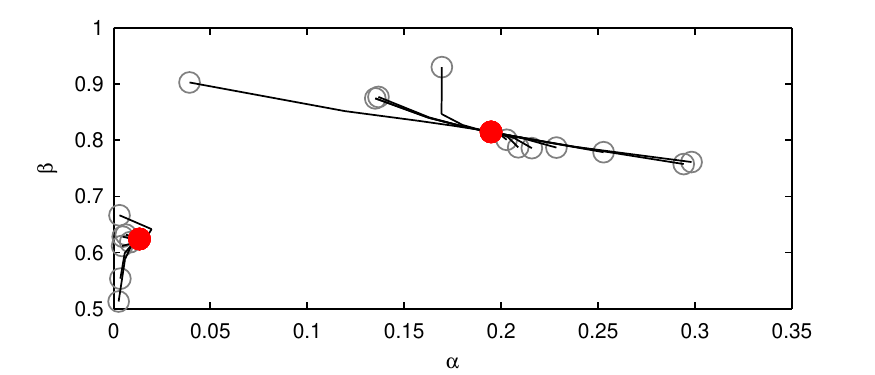}
 \caption{Segmentation results for power traces from the Pecan street data set
   with refrigerator on left and A/C on right. We model each device with an AR(1) process
   parameterized as $y_t = \alpha_t + \beta_ty_{t-1}$ and in the middle row show
   a particular segmentation for each device using $\lambda= 0.1$ and $1$,
   respectively. The bottom row shows the result of kernel density
   estimation for clustering the parameter space, identifying 2 modes (off/on) for the A/C and 3
   modes for the refrigerator---the third capturing the initial spike when the
   refrigerator switches on.}
 \label{fig-energy}
 \end{figure}

{\bf Modeling sources of energy consumption.} In our next example, we consider
the task of modeling energy consumed by household appliances using data from
Pecan Street, Inc. (\url{http://www.pecanstreet.org/}) collected with current
sensors installed at the circuit level inside the home. In this data set each
device has a unique energy profile and our goal is to build accurate models
which can be used to understand energy consumption in order to improve
energy efficiency. In Figure \ref{fig-energy} (top) we show typical power traces for
two such devices, a refrigerator and A/C unit---these devices that are characterized by a
small number of states and and their energy usage demonstrates strong
persistence between being on/off which we capture with an AR(1) model
parameterized by $y_t = \alpha_t + \beta_ty_{t-1}$. Empirically (as in the
previous example) we found that the probabilistic approach
improved significantly upon the simpler linear regression model which
typically did not produce segments corresponding to logical device states (the on
state was often over-segmented). In contrast, Figure
\ref{fig-energy} (bottom) shows the estimated modes from probabilistic
segmentation which correspond to an off state, on state, and in the case of the
refrigerator, a state representing the initial spike in energy consumed when the device
transitions from off to on.

\begin{figure}[t]
\includegraphics{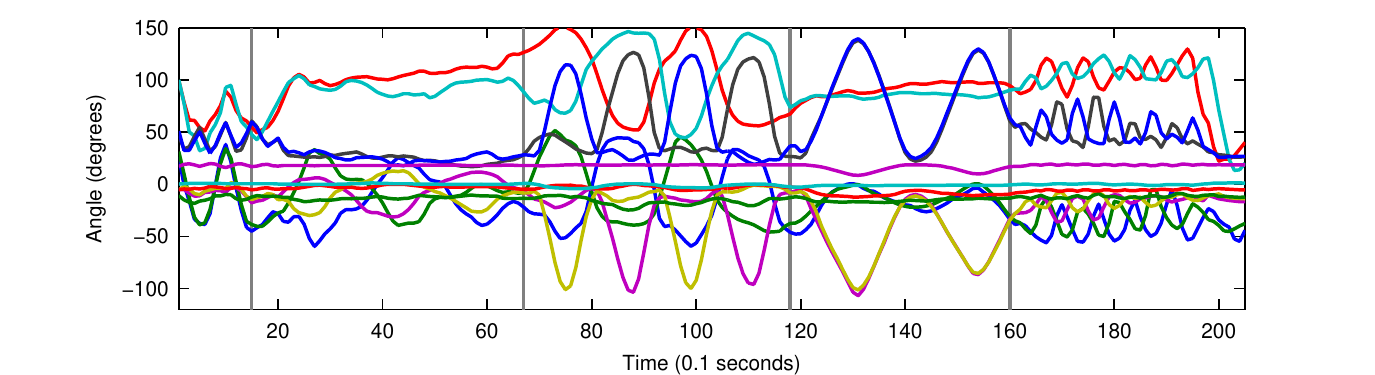}
\includegraphics{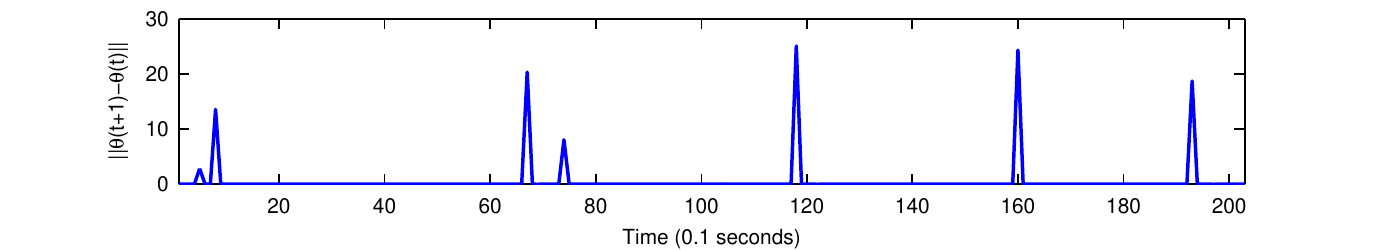}
\includegraphics[width=1.8in,height=1.8in]{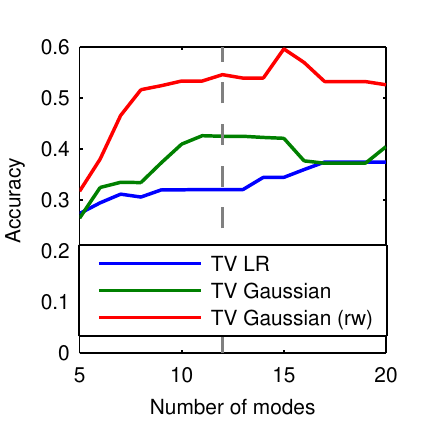}
\includegraphics{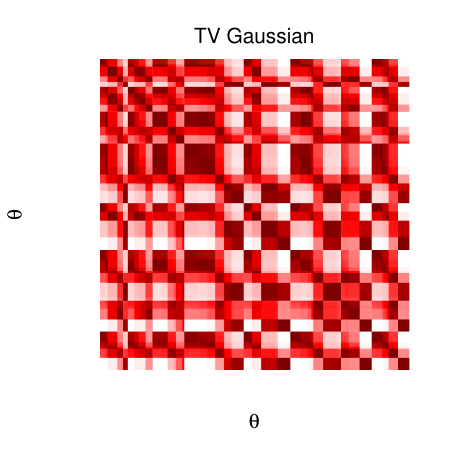}
\includegraphics{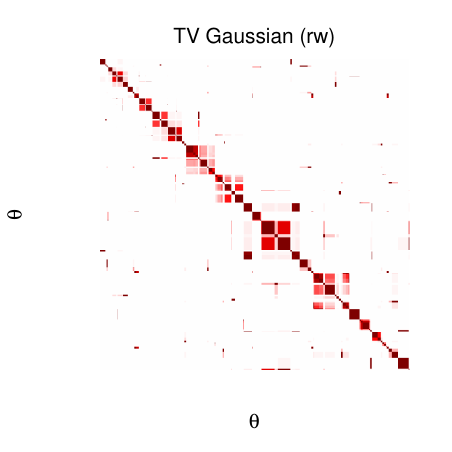}
\caption{Segmentation results on motion capture with input data of
  12 joint angles measured at 10Hz and manual segmentation on top; in middle,
  the segmentation from the AR(2) Gaussian
  model plus iterative reweighting. On bottom, we
  compare accuracy on identifying actions (segmentation plus clustering) with
  the reweighted Gaussian model performing the best. We examine this behavior on bottom
  middle/right, by reordering  $\theta_1,\ldots,\theta_T$ so that identical actions
  are adjacent and consider distance between pairs  $(\theta_t,\theta_{t'})$ with dark
  signifying closer. Comparing middle (non-reweighted) vs. right (reweighted),
  the evident block structure indicates that parameters for the same action are
  relatively closer after reweighting (presumably) allows for better parameter estimation.}
\label{fig-mocap}
\end{figure}

 {\bf Motion capture.} In our final example we consider segmenting motion capture
data, a task first proposed in \cite{fox2013joint} in conjunction with a
hierarchical nonparametric Bayesian model specifically designed to jointly model
behavior across subjects. We attempt to replicate that experimental setup here
 which includes sub-selecting from 62 available measurements a representative set of 12 angles
characterizing the behavior of the subject and manually labeling the sequences
with one of 12 actions. A typical sequence is shown in
Figure \ref{fig-mocap} (top), which (after normalization) we take as the output
variables $y_t \in \mathbb{R}^{12}$. As the signal shows not only persistence
but also clear periodic structure, we model this as an AR(2) process resulting
in $x_t \in \mathbb{R}^{25}$ and a parameter space with much higher dimension
that in previous examples (in the Gaussian model, $\Lambda_t \in
\mathbb{R}^{12 \times 12}$ and $\Theta_t \in \mathbb{R}^{25 \times 12}$).

First, in Figure \ref{fig-mocap} (middle) we have an accurate segmentation
provided by the Gaussian model  with the
additional step of a few iterations of iterative reweighting
\cite{candes2008enhancing}, an extension to the algorithm that previous authors
have suggested in the case of linear regression
\cite{ohlsson2010segmentation}. We found that while all methods
considered on this data set performed well on segmentation, the addition of the reweighting
step improved parameter estimation significantly resulting in the better
performance shown in Figure \ref{fig-mocap} (bottom left); here the comparison
depends on both accurate segmentation and parameter estimation in order
for density-based clustering to identifying similar segments which is required
to do well on this task.

The intuition behind the improvement from reweighting is shown in
Figure \ref{fig-mocap} (bottom right) which compares the distance of the
parameters after segmentation for the Gaussian model and the Gaussian model plus
reweighting---we see that when the parameters are allowed to
vary more significantly between segments (as a consequence of reweighting), the
parameters corresponding to the same action remains close
relative to the parameters for different actions. This is likely due to
better parameter estimation on the individual segments
from reducing the bias from total variation regularization. Overall, the
reweighted Gaussian model achieves accuracy of around 60\% which is comparable
to most previous results from \cite{fox2013joint} but somewhat worse than the
best model which is specifically designed for this task and benefits from highly
structured prior information.

\section{Conclusions and future work}

At a basic level, the techniques proposed in this paper center around
finding a convex (and hence, local-optima-free) approach to modeling
time series data in a manner that naturally segments the data into
different probabilistic models.  While the proposed method works well
in many settings, numerous extensions are possible in the overall
framework.  For instance, we can consider imposing additional
regularization on the joint space of parameters to enforce further
structure; we can generalize the getting to many other possible loss
functions; we can generalize the total variation penalty to the more
general $\ell_1$ trend filtering setting \cite{kim2009ell_1}, to
capture linear or higher order smooth segments in parameter space; and
we can extend the total variation penalty non-adjacent time points,
potentially directly allowing for segmentation across non-contiguous
regions. Further, we can explore extensions to kernel density
estimation in the parameter space that explicitly model the evolution
of these parameters, allowing us to build a full generative model rather than
just finding  modes as we do now.  Together, we believe that this
combination of approaches can lead to time series methods competitive
with latent variable models in terms of their flexibility and
representational power, but which are substantially easier and more
efficient to build and learn.

\bibliographystyle{plain}
\bibliography{convex_seg}

\begin{thebibliography}{10}

\bibitem{alaiz2013group}
Carlos~M Ala{\i}z, {\'A}lvaro Barbero, and Jos{\'e}~R Dorronsoro.
\newblock Group fused lasso.
\newblock {\em Artificial Neural Networks and Machine Learning--ICANN 2013},
  page~66, 2013.

\bibitem{basseville1993detection}
Mich{\`e}le Basseville, Igor~V Nikiforov, et~al.
\newblock {\em Detection of abrupt changes: theory and application}, volume
  104.
\newblock Prentice Hall Englewood Cliffs, 1993.

\bibitem{bleakley2011group}
Kevin Bleakley and Jean-Philippe Vert.
\newblock The group fused lasso for multiple change-point detection.
\newblock {\em arXiv preprint arXiv:1106.4199}, 2011.

\bibitem{boyd2011distributed}
Stephen Boyd, Neal Parikh, Eric Chu, Borja Peleato, and Jonathan Eckstein.
\newblock Distributed optimization and statistical learning via the alternating
  direction method of multipliers.
\newblock {\em Foundations and Trends{\textregistered} in Machine Learning},
  3(1):1--122, 2011.

\bibitem{candes2008enhancing}
Emmanuel~J Candes, Michael~B Wakin, and Stephen~P Boyd.
\newblock Enhancing sparsity by reweighted ℓ 1 minimization.
\newblock {\em Journal of Fourier analysis and applications}, 14(5-6):877--905,
  2008.

\bibitem{fox2011bayesian}
Emily Fox, Erik~B Sudderth, Michael~I Jordan, and Alan Willsky.
\newblock Bayesian nonparametric inference of switching dynamic linear models.
\newblock {\em Signal Processing, IEEE Transactions on}, 59(4):1569--1585,
  2011.

\bibitem{fox2013joint}
Emily~B Fox, Michael~C Hughes, Erik~B Sudderth, and Michael~I Jordan.
\newblock Joint modeling of multiple time series via the beta process with
  application to motion capture segmentation.
\newblock {\em arXiv preprint arXiv:1308.4747}, 2013.

\bibitem{fox2011sticky}
Emily~B Fox, Erik~B Sudderth, Michael~I Jordan, Alan~S Willsky, et~al.
\newblock A sticky hdp-hmm with application to speaker diarization.
\newblock {\em The Annals of Applied Statistics}, 5(2A):1020--1056, 2011.

\bibitem{kim2009ell_1}
Seung-Jean Kim, Kwangmoo Koh, Stephen Boyd, and Dimitry Gorinevsky.
\newblock $ell\_1$ trend filtering.
\newblock {\em Siam Review}, 51(2):339--360, 2009.

\bibitem{magnus1988matrix}
Jan~R Magnus and Heinz Neudecker.
\newblock Matrix differential calculus with applications in statistics and
  econometrics.
\newblock 1988.

\bibitem{oh2008learning}
Sang~Min Oh, James~M Rehg, Tucker Balch, and Frank Dellaert.
\newblock Learning and inferring motion patterns using parametric segmental
  switching linear dynamic systems.
\newblock {\em International Journal of Computer Vision}, 77(1-3):103--124,
  2008.

\bibitem{ohlsson2013identification}
Henrik Ohlsson and Lennart Ljung.
\newblock Identification of switched linear regression models using
  sum-of-norms regularization.
\newblock {\em Automatica}, 49(4):1045--1050, 2013.

\bibitem{ohlsson2010segmentation}
Henrik Ohlsson, Lennart Ljung, and Stephen Boyd.
\newblock Segmentation of arx-models using sum-of-norms regularization.
\newblock {\em Automatica}, 46(6):1107--1111, 2010.

\bibitem{rabiner1989tutorial}
Lawrence Rabiner.
\newblock A tutorial on hidden markov models and selected applications in
  speech recognition.
\newblock {\em Proceedings of the IEEE}, 77(2):257--286, 1989.

\bibitem{recht2010guaranteed}
Benjamin Recht, Maryam Fazel, and Pablo~A Parrilo.
\newblock Guaranteed minimum-rank solutions of linear matrix equations via
  nuclear norm minimization.
\newblock {\em SIAM review}, 52(3):471--501, 2010.

\bibitem{silverman1981using}
Bernard~W Silverman.
\newblock Using kernel density estimates to investigate multimodality.
\newblock {\em Journal of the Royal Statistical Society. Series B
  (Methodological)}, pages 97--99, 1981.

\bibitem{sun2006switched}
Zhendong Sun.
\newblock {\em Switched linear systems: Control and design}.
\newblock Springer, 2006.

\bibitem{wahlberg2012admm}
Bo~Wahlberg, Stephen Boyd, Mariette Annergren, and Yang Wang.
\newblock An admm algorithm for a class of total variation regularized
  estimation problems.
\newblock {\em arXiv preprint arXiv:1203.1828}, 2012.

\bibitem{wytock2014fast}
Matt Wytock, Suvrit Sra, and J.~Zico Kolter.
\newblock Fast {N}ewton methods for the group fused lasso.
\newblock In {\em Uncertainty in Artificial Intelligence}, 2014.

\bibitem{xuan2007modeling}
Xiang Xuan and Kevin Murphy.
\newblock Modeling changing dependency structure in multivariate time series.
\newblock In {\em Proceedings of the 24th international conference on Machine
  learning}, pages 1055--1062. ACM, 2007.

\bibitem{yuan2006model}
Ming Yuan and Yi~Lin.
\newblock Model selection and estimation in regression with grouped variables.
\newblock {\em Journal of the Royal Statistical Society: Series B (Statistical
  Methodology)}, 68(1):49--67, 2006.

\end{thebibliography}

\end{document}